\documentclass[11pt,a4paper]{article}
\usepackage[hyperref]{emnlp-ijcnlp-2019}
\usepackage{times}
\usepackage{latexsym}
\usepackage{graphicx}
\usepackage{booktabs,multirow}
\usepackage{caption}
\usepackage[whole]{bxcjkjatype}
\usepackage{bbding}
\usepackage{pifont}
\usepackage[nointegrals]{wasysym}
\usepackage{amssymb}
\usepackage{pifont}
\usepackage{tabularx}
\usepackage{listings}
\usepackage{amsmath,amssymb,mathtools}
\usepackage{bm}

\usepackage{url}
\usepackage{xspace}

\usepackage{url}

\aclfinalcopy 

\title{TEASPN: Framework and Protocol \\ for Integrated Writing Assistance Environments}

\author{Masato Hagiwara\textsuperscript{1} \hspace{6pt}
        Takumi Ito\textsuperscript{2,3} \hspace{6pt}
        Tatsuki Kuribayashi\textsuperscript{2,3} \hspace{6pt}
        Jun Suzuki\textsuperscript{2,4} \hspace{6pt}
        Kentaro Inui\textsuperscript{2,4} \\
  \textsuperscript{1}Octanove Labs LLC \hspace{12pt}
  \textsuperscript{2}Tohoku University \hspace{12pt}
  \textsuperscript{3}Langsmith Inc. \hspace{12pt}
  \textsuperscript{4}RIKEN AIP
  \\
  {\tt masato@octanove.com} \\
  {\tt \{t-ito, kuribayashi, jun.suzuki, inui\}@ecei.tohoku.ac.jp}
\\}

\date{}

\begin{document}
\maketitle
\begin{abstract}
Language technologies play a key role in assisting people with their writing. Although there has been steady progress in e.g., grammatical error correction (GEC), human writers are yet to benefit from this progress due to the high development cost of integrating with writing software. We propose TEASPN\footnote{See \url{https://www.teaspn.org/demo} for the screencast and \url{https://www.teaspn.org/} for more general info about TEASPN.}, a protocol and an open-source framework for achieving integrated writing assistance environments. The protocol standardizes the way writing software communicates with servers that implement such technologies, allowing developers and researchers to integrate the latest developments in natural language processing (NLP) with low cost. As a result, users can enjoy the integrated experience in their favorite writing software. The results from experiments with human participants show that users use a wide range of technologies and rate their writing experience favorably, allowing them to write more fluent text.
\end{abstract}

\section{Introduction}

Language technologies have been playing an important role in assisting people in writing natural language texts, such as essays, emails, business documents, and academic papers. There has been considerable progress on writing assistance technologies (or {\it WATs} in short) in the past few decades in fields such as NLP and computer-aided language learning (CALL). For example, in one of such areas, grammatical error correction (GEC)~\cite{Leacock:2010}, new models and systems are developed and published month after month, breaking the previous evaluation records and advancing state of the art. The recent development in neural language models enabled the completion of a prompt with long, realistic looking yet coherent passages~\cite{Radford:2019}.

{\setlength\textfloatsep{0pt}
\begin{figure}[t]
    \centering
      \includegraphics[width=5.5cm]{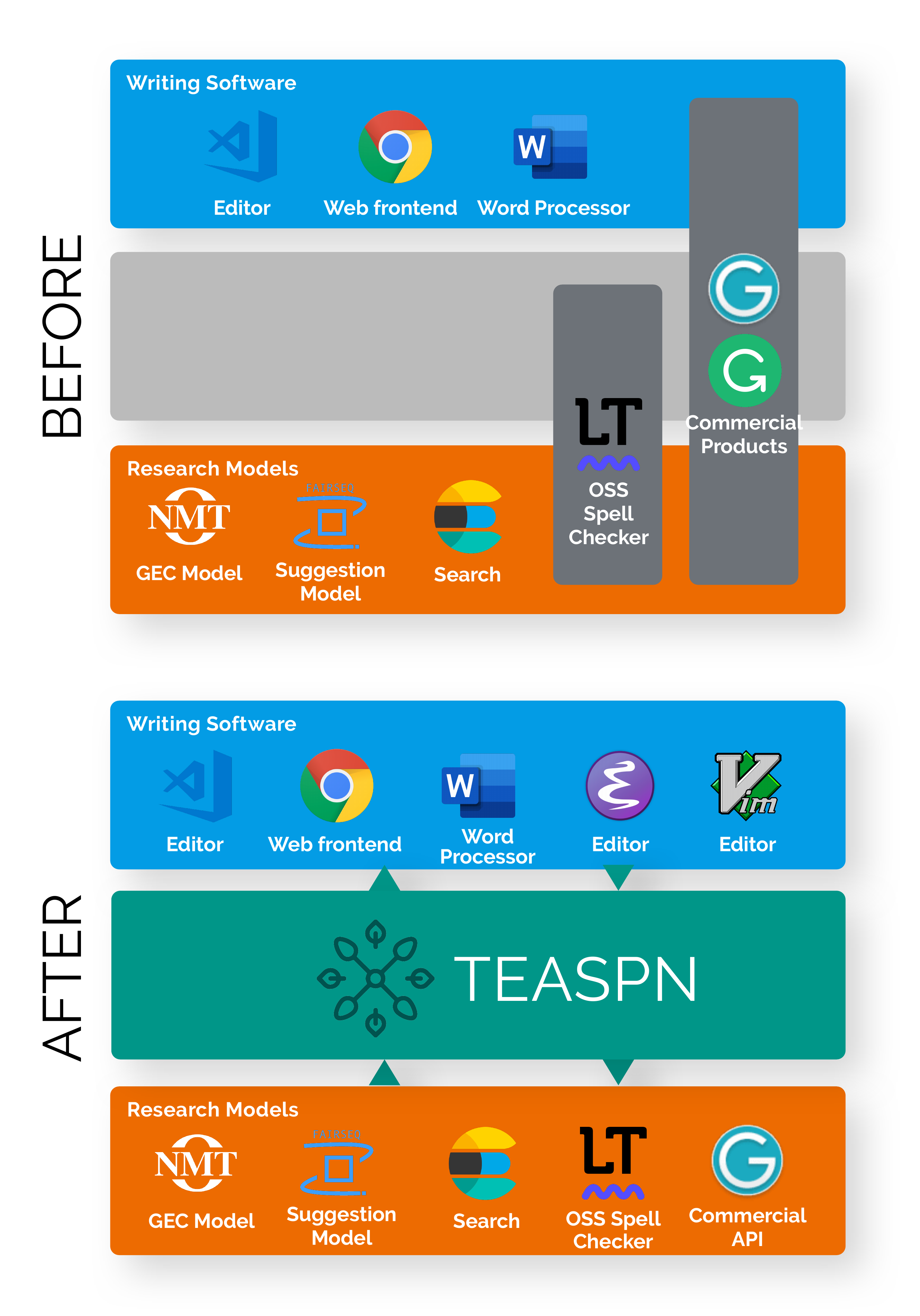}
      \vspace{-0.4cm}
      \caption{Writing software before and after TEASPN.}
      \vspace{-0.2cm}
      \label{fig:overview}
\end{figure}
}

However, real-world users such as writers who can and should benefit the most from WATs are yet to reap the fruits from these research efforts. Aside from a small number of commercial products, notably Grammarly\footnote{\url{https://www.grammarly.com/}} and Smart Compose~\cite{Chen:2019}, and research systems such as WriteAhead~\cite{Yen:2015,Chang:2015} and CroVeWA~\cite{Soyer:2015}, we see few examples of user-facing applications and experiments that make use of recent development in WATs. Many models are confined in research implementations that are not easily accessible to end users and the larger society. WATs, however, are not truly useful until they are integrated into user-facing writing applications such as editors and word processors (collectively called {\it writing software} in this paper) and interact with end users in a dynamic and intuitive manner. This ``great divide'' (see Figure \ref{fig:overview} BEFORE) between applications and academia is not unique in the domain of writing assistance, but a widespread phenomenon across many fields in machine learning and NLP, as pointed out by \citet{Wagstaff:2012}.

One cause of this ``great divide'' is the high development cost for integrating and bridging both sides. Since there is a wide range of WATs, it is impractical, if not impossible, for developers of writing software to support all types of such technologies that come in different packages in different programming languages. Similarly, since there is a large selection of writing software, WAT researchers and developers cannot afford to offer their solutions in such a way that most writing software packages can benefit from them. If there are $N$ types of writing software and $M$ types of WATs, there can be $N \times M$ combinations between the two sides. As a result, writers often need to rely on many different writing software solutions and switch between many different applications and websites (search engines, grammar checkers, dictionaries and thesauri, etc.) in order to complete their tasks.

In this paper, we propose TEASPN (Text Editing Assistance Smartness Protocol for Natural Language; pronounced ``teaspoon''), a protocol and a framework for achieving integrated writing assistance environments, as a solution to this ``great divide'' problem (Figure \ref{fig:overview} AFTER). Inspired by and built upon Language Server Protocol (LSP)\footnote{\url{https://microsoft.github.io/language-server-protocol/}}, a similar protocol for integrating software development environments, TEASPN provides an open protocol that standardizes the way writing software and WATs communicate with each other. We also released the TEASPN SDK (software development kit) as an open source library, which eases the cost of making WATs compatible with TEASPN. As a result, by using TEASPN,

\begin{itemize}
    \setlength{\parskip}{0cm}
    \setlength{\itemsep}{0cm}
    \item Developers of writing software can easily integrate state-of-the-art WATs into their editors and word processors just by following the protocol.
    \item Developers and researchers of WATs can support major writing software applications without worrying about the development cost, just by using the TEASPN SDK.
    \item Writers can benefit from integrated writing experience provided by their favorite writing software and WATs.
\end{itemize}

Finally, we implemented a demo TEASPN server that integrates WATs using latest developments in NLP (e.g., a neural language model and seq2seq-based paraphrasing) and ran experiments with real human writers to verify the framework's effectiveness. The experimental results demonstrated that our integrated writing assistance system developed with TEASPN provides better writing experience for human writers.

\section{Related Work}

\paragraph{Writing assistance} Use of language technologies for assisting writing in a second language (L2) has been extensively explored, especially for non-native English speakers. One of the most active research areas is GEC~\cite{Leacock:2010}, where several new models are published every year and commercial systems such as Grammarly are actively developed. Other research-based systems include WriteAhead~\cite{Yen:2015,Chang:2015}, an interactive writing environment that provides users with grammatical patterns mined from large corpora, and CroVeWA~\cite{Soyer:2015}, a crosslingual sentence search system for L2 writers. FLOW~\cite{Chen:2012} is another writing assistance system that allows users to type in their first languages (L1) and suggests words and phrases in L2. Running syntactic analysis and visualizing sentence structures have also been explored for L2 reading assistance~\cite{Faltin:2003,Srdanovic:2011}.

In addition to L2 learners, the use of technologies for assisting human translators has also been a focus of research. TransType~\cite{Langlais:2000} is a translation assistance system that suggests completions for the text to the human translator in an interactive manner. In SemEval 2014, \citet{vanGompel:2014l} presented an L2 writing assistance task where systems find the proper translation of a word given a context in L2. Other writing assistance systems (not necessarily L2 learners) include assisting users with composing an email by auto-completion~\cite{Chen:2019} and reply suggestion~\cite{Kannan:2016}.

\paragraph{LSP} The $N\times M$ problem mentioned in Section~1 is not unique to writing assistance. In software development, there can be $N$ different types of integrated development environments (IDEs) and $M$ different programming languages, making the integration cost proportionally expensive to $N \times M$. LSP solved this problem by proposing an open protocol that standardizes the way IDEs communicate with servers that offer language smartness technologies such as syntax checking and completion. As of today, LSP is widely adopted and supported by more than 70 servers and 20 development environments.

Since there is a large overlap between authoring in programming and natural languages, we built TEASPN as a ``fork'' of LSP. There are a few features that we need to design and implement specifically for writing assistance, namely, syntax highlighting and external resource search, which makes TEASPN incompatible with LSP. However, we re-purposed many LSP data models and features for TEASPN. Being able to leverage existing resources for LSP gives TEASPN a great head start for a wide adoption.

\paragraph{Protocols for NLP} Language Grid~\cite{Ishida:2006} is a platform where language providers (e.g., translation systems) and linguistic resources (e.g., dictionaries) are connected via semantic Web technologies to provide language services to communities. The NLP Interchange Format (NIF)~\cite{Hellmann:2012} is a standard that aims to achieve interoperability between different NLP tools and resources by defining an RDL/OWL-based format. Although these projects have seen some real-world success, their adoption is quite limited as of this writing, compared to the aforementioned LSP, which powers at least a couple of millions of developers worldwide both for Visual Studio Code (VS code)\footnote{\url{https://code.visualstudio.com/blogs/2017/11/16/connect}} and Atom\footnote{\url{https://blog.atom.io/2016/03/28/atom-reaches-1m-users.html}}. We believe the key to the wide adoption of any protocol is the focus on the right scope, practicality, and ease of development, which are the guiding principles for TEASPN.

\section{TEASPN}

{\setlength\textfloatsep{0pt}
\begin{figure}[t]
    \centering
      \includegraphics[height=7.0cm]{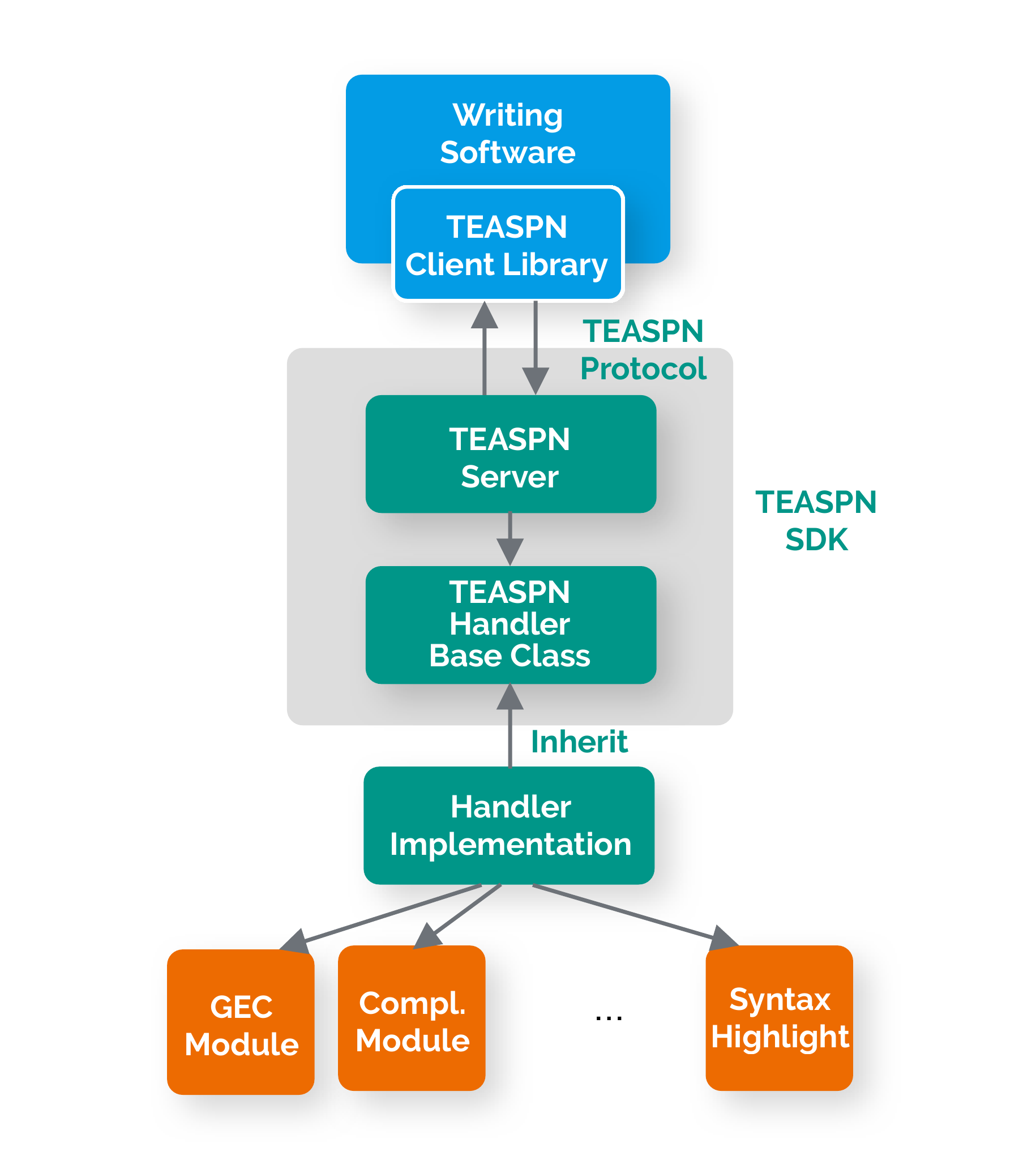}
      \vspace{-0.1cm}
      \caption{Architecture of TEASPN.}
      \label{fig:architecture}
      \vspace{-0.1cm}
\end{figure}
}

\subsection{Overview}

TEASPN adopts a client-server architecture (Figure \ref{fig:architecture}), where a client (writing software such as an editor or a word processor) communicates with a server that provides WATs. The client and the server communicate over the TEASPN protocol, an HTTP-like protocol which uses JSON-RPC (remote procedure call)\footnote{\url{http://www.jsonrpc.org/}} to encode the message body. Requests can be sent in both directions, which are often triggered by some events (such as user input) and can be responded with additional data (such as results of GEC). TEASPN clients and servers can be written in any programming language as long as they conform to the protocol.

\subsection{Features}

By integrating a large selection of WATs in a single platform, TEASPN makes them available to writers at their fingertips and is expected to improve writing effectiveness. Table \ref{tab:features} shows the list of WATs that are supported by TEASPN. Notice that the list includes a wide range of WATs that have been extensively investigated the literature (e.g., GEC and search) as well as the ones that are less explored (e.g., syntax highlighting and jump). 

Although we were able to build many WAT features upon existing ones from LSP, there were two features that we needed to design from scratch specifically for TEASPN—syntax highlighting and search. While syntax highlighting is usually handled on the client side for programming languages using shallow lexical analysis, syntactic analyses of natural language can be too costly and complex to be handled solely by the client. Therefore, we defined a new type of request and related type declarations so that it can be handled by the server.

The second feature, search, enables writers to search external linguistic resources. While the search feature for LSP is limited to the files in the same workspace, writers of natural language texts often need to consult a wide variety of resources such as corpora and dictionaries. 

\begin{table*}[]
\centering
\small
\begin{tabular}{ll}
\toprule
Feature                      & Description \\
\cmidrule(r){1-1} \cmidrule(l){2-2}
Syntax highlighting          & Highlighting parts of text \\
Grammatical error detection (GED) & Detecting typological and grammatical errors  \\
Grammatical error correction (GEC) & Automatically correcting issues detected by GED  \\
Completion                   & Completing or suggest succeeding text \\
Text rewriting               & Rewriting part of text (paraphrasing, translation, etc.) \\
Jump                         & Jumping to other locations (coreference, definitions, etc.) \\
Hover                        & Showing extra information about the location (e.g., definition) \\
Search                       & Searching external resources such as corpora and dictionaries \\ \bottomrule
\end{tabular}
 \caption{WATs supported by TEASPN.}
 \label{tab:features}
\end{table*}

\subsection{Developers of Writing Software}
\label{subsec:software}

By adopting TEASPN, developers of writing software can easily integrate WATs into their editors and word processors. The fact that a large number of IDEs and editors already support LSP helps to a great extent. For example, in the sample TEASPN client implementation\footnote{\url{https://github.com/teaspn/teaspn-sdk}} we provide for VS Code, we needed to modify less than 200 lines of TypeScript code to make it compatible with the TEASPN protocol while leveraging the existing library for LSP. We were also able to implement preliminary TEASPN clients for Atom and Sublime text\footnote{\url{https://www.sublimetext.com/}} with little modification to existing code.

\subsection{Developers of WATs}
\label{subsec:technologies}

Developers and researchers of WATs can make their technologies available to major writing software just by using TEASPN without writing any client code. To facilitate the development process, we released the TEASPN SDK, which includes a library and a sample TEASPN server implementation in Python, one of the most popular programming languages for developing language technologies as of late. The library takes care of low-level communication and text synchronization with the client, letting WAT developers just inherit the TEASPN handler base class and focus on implementing the missing core NLP logic. As an example, Figure \ref{fig:snippet} shows a simplified code snippet for implementing completion. Notice the brevity of the code. We also provide a simple yet working TEASPN server implementation in the SDK for reference to accelerate the development.

{\setlength\textfloatsep{0pt}
\begin{figure}[t]
    \centering
      \includegraphics[height=3cm]{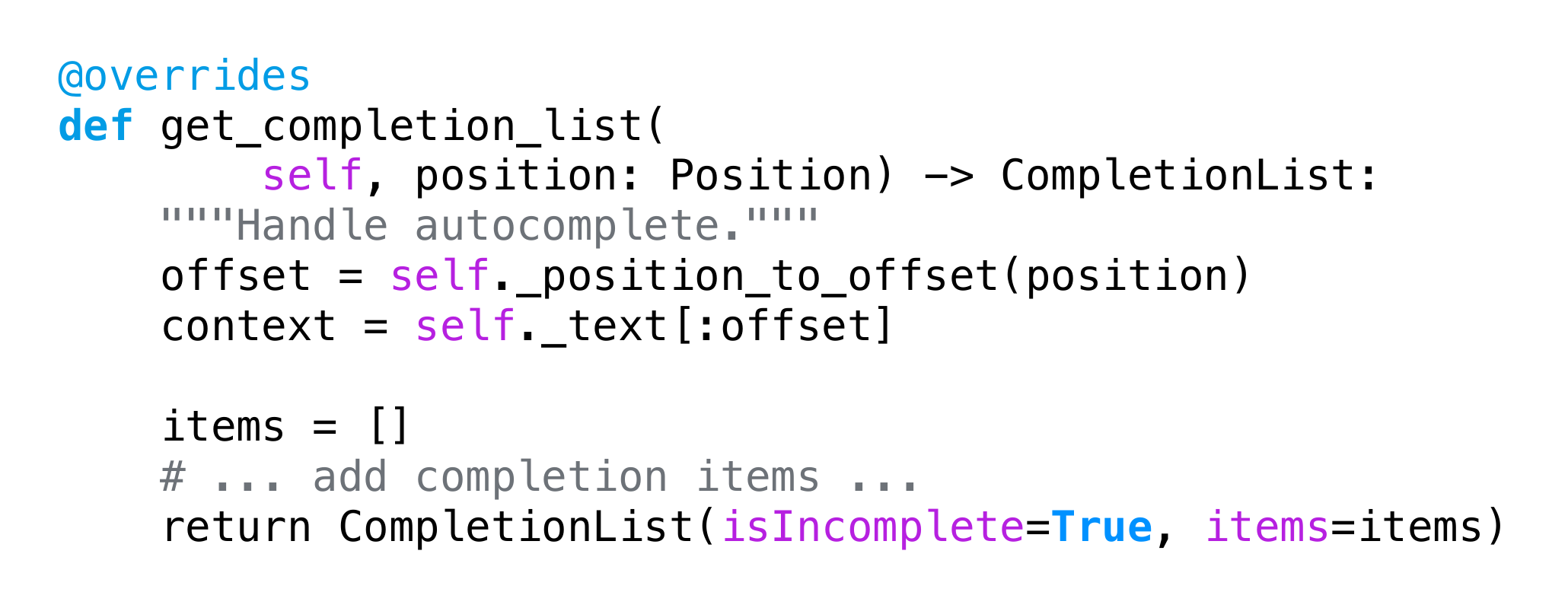}
      \vspace{-0.5cm}
      \caption{Code snippet for computing completion}
      \label{fig:snippet}
\end{figure}
}

\section{Experiments and Implementation}
\subsection{Experiments}

In this section, we develop a demo TEASPN system and investigate its effectiveness through experiments with end-users (writers) in order to answer the following research question:

\textit{Does the integrated writing assistance environment developed with TEASPN provide better writing experience and help write better texts?}

To explore the effectiveness of integrated writing assistance environment, we compared the following two conditions. In the {\bf INTEGRATED} condition, participants used an editor (VS Code) equipped with TEASPN, where many WATs were available, while in the {\bf BASELINE} condition, they used the same editor with no WATs activated, while being allowed to use any other writing tools outside the editor (e.g., Grammarly and Web dictionaries). The BASELINE condition was set up to simulate the real situation that writers face, where writing assisting technologies are implemented separately outside the editor.

The participants of our experiments consist of twelve college students or researchers in NLP with a diverse L1 distribution: Bengali: 1, Chinese: 1, Croatian: 1, German: 1, Hindi: 1, Japanese: 6, Spanish: 1. They were directed to go through two writing sessions, one for each condition mentioned above, during which they wrote English text within five sentences in response to two different prompts: (i) {\it write about an activity you enjoy, such as a hobby}, and (ii) {\it write about your hometown}. The prompts and the writing environments were combined randomly. Before the writing sessions came an instruction session, where the authors of this paper showed the participants all the features of the demo system for a demonstration purpose. After the writing sessions, they were asked questions regarding their writing experience. Because the focus of this experiment is to evaluate the integrated writing assistance environment, participants are instructed not to consider the performance of individual WATs.

\subsection{Implementation of the Demo System}

{\setlength\textfloatsep{4pt}
\begin{figure}[t]
    \centering
      \includegraphics[height=3.2cm]{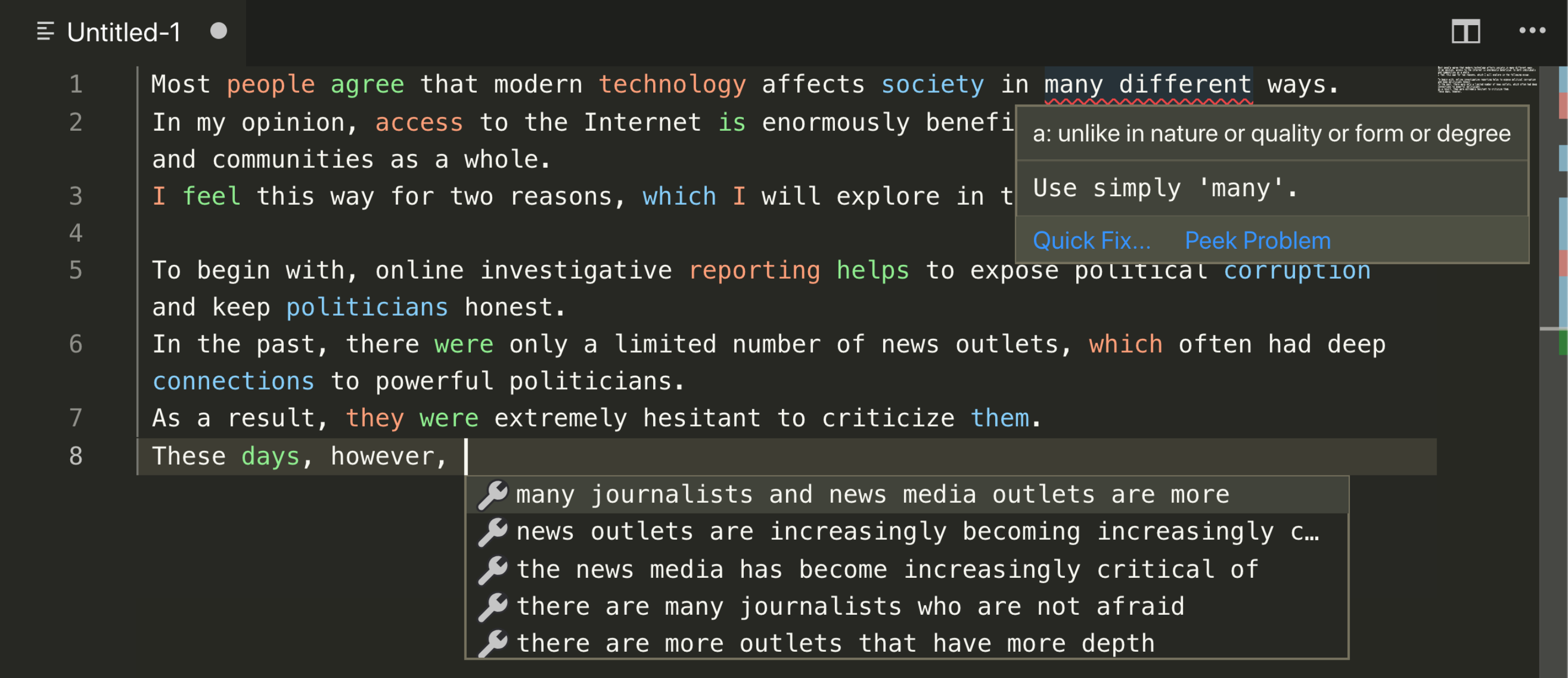}
      \caption{Screenshot of our demo system.}
      \label{fig:demo}
\end{figure}
}

We implemented a demo system using the TEASPN framework which has all of the features shown in Table~\ref{tab:features}. 
See Figure~\ref{fig:demo} for the screenshot.

For syntax highlighting, we used the dependency parser  SpaCy\footnote{\url{https://spacy.io/}}.
Head tokens with specific dependency relation\footnote{{\tt ROOT}, {\tt nsubj}, {\tt nsubjpass}, and {\tt dobj} in the CLEAR style tag set.} were highlighted in different colors.
As for the GEC and GED features, we used the open-source GEC tool LanguageTool 3.2\footnote{\url{https://github.com/languagetool-org/languagetool/releases/tag/v3.2}}.
We implemented two types of completion features: one which suggests the likely next phrases given the context using a neural language model~\citep{Radford:2019} and the other one which suggests a set of words consistent with the characters being typed.
We built a seq2seq paraphrase model trained on PARANMT-50M~\citep{Wieting:2018} for the text rewriting feature, which allows the writer to select a part of the text and chooses among paraphrases.
As for the jump feature, we used a coreference resolution model\footnote{\url{https://github.com/huggingface/neuralcoref}} to jump from a selected expression to its antecedent.
The hover feature shows the definition of a hovered word using WordNet\footnote{\url{https://wordnet.princeton.edu/}}.
Finally, we implemented a full-text search feature using the open multilingual sentence dataset Tatoeba\footnote{\url{https://tatoeba.org/eng/}} and used Elasticsearch 7.1.1\footnote{\url{https://www.elastic.co}} for indexing and search.

\section{Results and Analysis}
{\setlength\textfloatsep{0pt}
\begin{figure}[t]
    \centering
      \includegraphics[width=7.5cm]{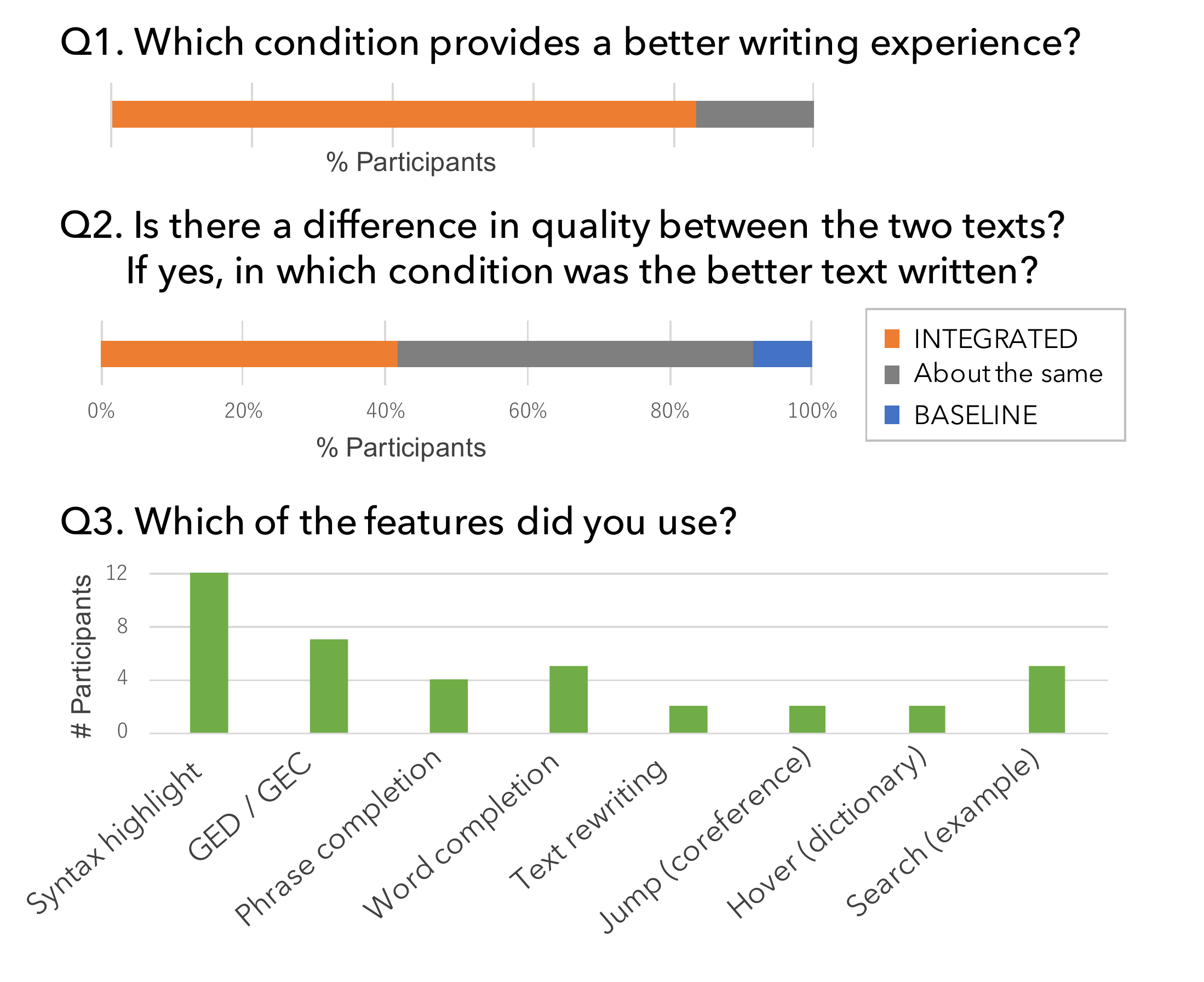}
      \vspace{-0.1cm}
      \caption{Response summary of the questionnaire.}
      \label{fig:results}
\end{figure}
}

After the writing sessions, the participants responded to a questionnaire including the following questions: {\it Q1: Which environment provides a better writing experience?}, {\it Q2: Is there a difference in quality between the two texts? If yes, in which environment was the better text written?}, and {\it Q3: Which of the following features did you use?}

Figure~\ref{fig:results} summarizes the responses from the participants. Ten out of twelve (83.3\%) participants rated their experience favorably (Q1), and 40\% believed they were able to write better texts in the INTEGRATED condition (Q2), demonstrating the effectiveness of the integrated writing assistance environment with TEASPN. 
The responses for Q3 show that an average participant used 3.2 WAT features in the INTEGRATED condition\footnote{Note that we assume that every participant used syntax highlighting, which is activated by default.}. This suggests that the writers can benefit from an integrated environment with various WATs activated.

\begin{table}[t]
    \centering
    \renewcommand{\arraystretch}{0.9}
    \small{
    \begin{tabular}{llll}
    \toprule
    Condition & Perplexity & \# Chars. (mean $\pm$ std) \\
    \cmidrule(r){1-1} \cmidrule(l){2-2} \cmidrule(l){3-3}
    BASELINE & 37.8 & 379 $\pm$ 116 \\ 
    \cmidrule(r){1-1} \cmidrule(l){2-2} \cmidrule(l){3-3}
    INTEGRATED & 26.4 & 335 $\pm$ 91 \\
    \bottomrule
    \end{tabular}
    }
 \caption{Statistics of the written texts.}
 \label{tab:analysis}
\end{table}

We ran further analyses on the texts written by the participants during the writing sessions. Table~\ref{tab:analysis} shows some statistics of the written texts in the two experimental conditions. Perplexity was calculated using the pretrained GPT-2 model (small, 117M parameters)~\citep{Radford:2019}.
The texts written in the INTEGRATED condition had lower perplexity, suggesting that the integrated writing environment helped them write more fluent and/or typical English text. This result backs up the subjective responses from the participants indicating they were able to produce better texts in the INTEGRATED condition than the other. We also note that the texts written in the INTEGRATED condition were relatively shorter. This could be due to the fact that the participants were still spending some of their time observing and figuring out the behavior of the assisting features and spending slightly less time actually writing. We believe this trend will disappear or even reverse itself as they get more used to the integrated writing experience and the quality of the inidividual WATs improve.

\section{Conclusion and Future Work}
We proposed TEASPN, a framework and a protocol which standardizes the way writing software communicates with writing assistance technologies, to achieve integrated writing assistance environments. In addition, we released the TEASPN SDK as an open source library, which eases the cost of making WATs compatible with TEASPN. We developed a demo system which implements various assistance technologies based on latest NLP developments and ran experiments with human participants. The result demonstrated that they rated their integrated writing experience favorably, potentially helping them write more fluent and better text.

In future work, by making this a larger community effort, we wish to broaden the support lineup for writing software while developing various writing assistance features with TEASPN, further closing the gap between the latest developments in NLP and real-world human users.

\bibliographystyle{acl_natbib}
\bibliography{emnlp-ijcnlp-2019}

\begin{thebibliography}{16}
\expandafter\ifx\csname natexlab\endcsname\relax\def\natexlab#1{#1}\fi

\bibitem[{Chang and Chang(2015)}]{Chang:2015}
Jim Chang and Jason Chang. 2015.
\newblock \href {https://doi.org/10.3115/v1/N15-3022} {{W}rite{A}head2: Mining
  lexical grammar patterns for assisted writing}.
\newblock In \emph{Proceedings of the {NAACL} 2015 System Demonstrations},
  pages 106--110.

\bibitem[{Chen et~al.(2012)Chen, Huang, Hsieh, Kao, and Chang}]{Chen:2012}
Mei-Hua Chen, Shih-Ting Huang, Hung-Ting Hsieh, Ting-Hui Kao, and Jason~S.
  Chang. 2012.
\newblock \href {https://www.aclweb.org/anthology/P12-3027} {{FLOW}: A
  first-language-oriented writing assistant system}.
\newblock In \emph{Proceedings of {ACL} 2012 System Demonstrations}, pages
  157--162.

\bibitem[{Chen et~al.(2019)Chen, Lee, Bansal, Cao, Zhang, Lu, Tsay, Wang, Dai,
  Chen, Sohn, and Wu}]{Chen:2019}
Mia~Xu Chen, Benjamin~N Lee, Gagan Bansal, Yuan Cao, Shuyuan Zhang, Justin Lu,
  Jackie Tsay, Yinan Wang, Andrew~M. Dai, Zhifeng Chen, Timothy Sohn, and
  Yonghui Wu. 2019.
\newblock \href {https://arxiv.org/pdf/1906.00080.pdf} {Gmail smart compose:
  Real-time assisted writing}.

\bibitem[{Faltin(2003)}]{Faltin:2003}
Anne~Vandeventer Faltin. 2003.
\newblock Natural language processing tools for computer assisted language
  learning.
\newblock \emph{Linguistik Online}, 17(5):137--153.

\bibitem[{van Gompel et~al.(2014)van Gompel, Hendrickx, van~den Bosch, Lefever,
  and Hoste}]{vanGompel:2014l}
Maarten van Gompel, Iris Hendrickx, Antal van~den Bosch, Els Lefever, and
  V{\'e}ronique Hoste. 2014.
\newblock \href {https://doi.org/10.3115/v1/S14-2005} {{S}em{E}val 2014 task 5
  - {L}2 writing assistant}.
\newblock In \emph{Proceedings of {S}em{E}val 2014}, pages 36--44.

\bibitem[{Hellmann et~al.(2012)Hellmann, Lehmann, and Auer}]{Hellmann:2012}
Sebastian Hellmann, Jens Lehmann, and Soren Auer. 2012.
\newblock {NIF}: An ontology-based and linked-data-aware {NLP} interchange
  format.
\newblock Technical report, Working Draft.

\bibitem[{Ishida(2006)}]{Ishida:2006}
Toru Ishida. 2006.
\newblock {L}anguage {G}rid: An infrastructure for intercultural collaboration.
\newblock In \emph{IEEE/IPSJ Symposium on Applications and the Internet (SAINT
  2006)}, pages 96--100.

\bibitem[{Kannan et~al.(2016)Kannan, Kurach, Ravi, Kaufmann, Tomkins, Miklos,
  Corrado, Luk{\'a}cs, Ganea, Young, and Ramavajjala}]{Kannan:2016}
Anjuli Kannan, Karol Kurach, Sujith Ravi, Tobias Kaufmann, Andrew Tomkins,
  Balint Miklos, Gregory~S. Corrado, L{\'a}szl{\'o} Luk{\'a}cs, Marina Ganea,
  Peter Young, and Vivek Ramavajjala. 2016.
\newblock Smart reply: Automated response suggestion for email.
\newblock In \emph{Proceedings of KDD 2016}.

\bibitem[{Langlais et~al.(2000)Langlais, Foster, and Lapalme}]{Langlais:2000}
Philippe Langlais, George Foster, and Guy Lapalme. 2000.
\newblock \href {https://www.aclweb.org/anthology/W00-0507} {{T}rans{T}ype: a
  computer-aided translation typing system}.
\newblock In \emph{{ANLP}-{NAACL} 2000 Workshop: Embedded Machine Translation
  Systems}.

\bibitem[{Leacock et~al.(2010)Leacock, Chodorow, Gamon, and
  Tetreault}]{Leacock:2010}
Claudia Leacock, Martin Chodorow, Michael Gamon, and Joel Tetreault. 2010.
\newblock \emph{Automated Grammatical Error Detection for Language Learners}.
\newblock Morgan \& Claypool.

\bibitem[{Radford et~al.(2019)Radford, Wu, Child, Luan, Amodei, and
  Sutskever}]{Radford:2019}
Alec Radford, Jeffrey Wu, Rewon Child, David Luan, Dario Amodei, and Ilya
  Sutskever. 2019.
\newblock Language models are unsupervised multitask learners.
\newblock Technical report, OpenAI.

\bibitem[{Soyer et~al.(2015)Soyer, Topi{\'c}, Stenetorp, and
  Aizawa}]{Soyer:2015}
Hubert Soyer, Goran Topi{\'c}, Pontus Stenetorp, and Akiko Aizawa. 2015.
\newblock \href {https://doi.org/10.3115/v1/N15-3019} {{C}ro{V}e{WA}:
  Crosslingual vector-based writing assistance}.
\newblock In \emph{Proceedings of {NAACL} 2015 System Demonstrations}, pages
  91--95, Denver, Colorado.

\bibitem[{Srdanović(2011)}]{Srdanovic:2011}
Irena Srdanović. 2011.
\newblock \href {https://www.aclweb.org/anthology/P12-3027} {Evaluating
  e-resources for {J}apanese language learning}.
\newblock In \emph{Proceedings of {eLex} 2011}, pages 260--267.

\bibitem[{Wagstaff(2012)}]{Wagstaff:2012}
Kiri Wagstaff. 2012.
\newblock Machine learning that matters.
\newblock In \emph{Proceedings of the Twenty-Ninth International Conference on
  Machine Learning (ICML)}, pages 529--536.

\bibitem[{Wieting and Gimpel(2018)}]{Wieting:2018}
John Wieting and Kevin Gimpel. 2018.
\newblock \href {https://www.aclweb.org/anthology/P18-1042} {{P}ara{NMT}-50{M}:
  Pushing the limits of paraphrastic sentence embeddings with millions of
  machine translations}.
\newblock In \emph{Proceedings of ACL 2018}, pages 451--462.

\bibitem[{Yen et~al.(2015)Yen, Wu, Chang, Boisson, and Chang}]{Yen:2015}
Tzu-Hsi Yen, Jian-Cheng Wu, Jim Chang, Joanne Boisson, and Jason Chang. 2015.
\newblock \href {https://doi.org/10.3115/v1/P15-4024} {{W}rite{A}head: Mining
  grammar patterns in corpora for assisted writing}.
\newblock In \emph{Proceedings of {ACL}-{IJCNLP} 2015 System Demonstrations},
  pages 139--144.

\end{thebibliography}

\appendix

\end{document}